\documentclass[10pt,twocolumn,letterpaper]{article}

\usepackage{times}
\usepackage{epsfig}
\usepackage{graphicx}
\usepackage{amsmath}
\usepackage{amssymb}
\usepackage{enumitem}

\usepackage{cite}                        
\usepackage{pifont}                      
\usepackage[usenames, dvipsnames, table]{xcolor} 
\usepackage{multicol}                    
\usepackage{multirow,makecell}           
\usepackage{booktabs}                    
\usepackage{adjustbox}                   
\usepackage{siunitx}                     
\usepackage{iccv}
\usepackage{subcaption}                  
\usepackage[font=small]{caption}         
\usepackage[pagebackref=true,breaklinks=true,letterpaper=true,colorlinks,bookmarks=false]{hyperref}

\usepackage[capitalise]{cleveref}

\iccvfinalcopy 
\def\iccvPaperID{21} 

\ificcvfinal\fi

\newcommand{\compactparagraph}[1]{\smallbreak\noindent\textbf{#1}}
\DeclareMathOperator{\T}{T}
\DeclareMathOperator*{\argmax}{arg\,max}

\newcommand{\Ty}{\T_y}
\newcommand{\Tyhat}{\hat{\T}_y}

\newcommand{\tick}{\checkmark}
\newcommand{\cross}{\ding{55}}

\definecolor{note}{RGB}{60, 173, 171}
\definecolor{todo-blue}{RGB}{79, 137, 228}
\definecolor{pink}{RGB}{243, 188, 255}
\definecolor{gray}{gray}{0.9}
\definecolor{yellow}{HTML}{FFCC00}
\definecolor{dark-red}{RGB}{120, 0, 0}
\definecolor{cell-blue}{HTML}{82C6E2}
\definecolor{sns-blue}{HTML}{1F77B4}
\definecolor{sns-purple}{HTML}{8D5FD3}

\definecolor{equivariant-color}{HTML}{1f77b4}
\definecolor{invariant-color}{HTML}{5b2ca2}

\definecolor{good}{HTML}{62c32f}
\definecolor{bad}{HTML}{c32f2f}

\newcommand{\class}[1]{`\textit{#1}'}
\newcommand{\jesterzs}[1]{Jester-#1}
\newcommand{\somethingzs}[1]{SS-#1}

\newcommand{\chl}[1]{\cellcolor{cell-blue!50} #1}
\newcommand{\cbhl}[1]{\cellcolor{red!20} #1}

\providecommand{\guideline}[1]{}
\providecommand{\todo}[1]{}
\providecommand{\dimaN}[1]{}
\providecommand{\willN}[1]{}

\begin{document}
\title{Retro-Actions: Learning `Close' by Time-Reversing `Open' Videos}

\author{Will Price\\
University of Bristol\\
{\tt\small will.price@bristol.ac.uk}
\and
Dima Damen\\
University of Bristol\\
{\tt\small dima.damen@bristol.ac.uk}
}

\maketitle
\thispagestyle{empty}

\begin{abstract}
We investigate video transforms that result in class-homogeneous label-transforms.
These are video transforms that consistently maintain or modify the labels of all videos in each class.
We propose a general approach to discover invariant classes, whose transformed examples maintain their label; pairs of equivariant classes, whose transformed examples exchange their labels; and novel-generating classes, whose transformed examples belong to a new class outside the dataset.
Label transforms offer additional supervision previously unexplored in video recognition benefiting data augmentation and enabling zero-shot learning opportunities by learning a class from transformed videos of its counterpart.

Amongst such video transforms, we study horizontal-flipping, time-reversal, and their composition.
We highlight errors in naively using horizontal-flipping as a form of data augmentation in video.
Next, we validate the realism of time-reversed videos through a human perception study where people exhibit equal preference for forward and time-reversed videos.
Finally, we test our approach on two datasets, Jester and Something-Something, evaluating the three video transforms for zero-shot learning and data augmentation.
Our results show that gestures such as \class{zooming in} can be learnt from \class{zooming out} in a zero-shot setting, as well as more complex actions with state transitions such as \class{digging something out of something} from \class{burying something in something}.
\end{abstract}

\begin{figure}[ht!]
  \centering \includegraphics[width=1\linewidth]{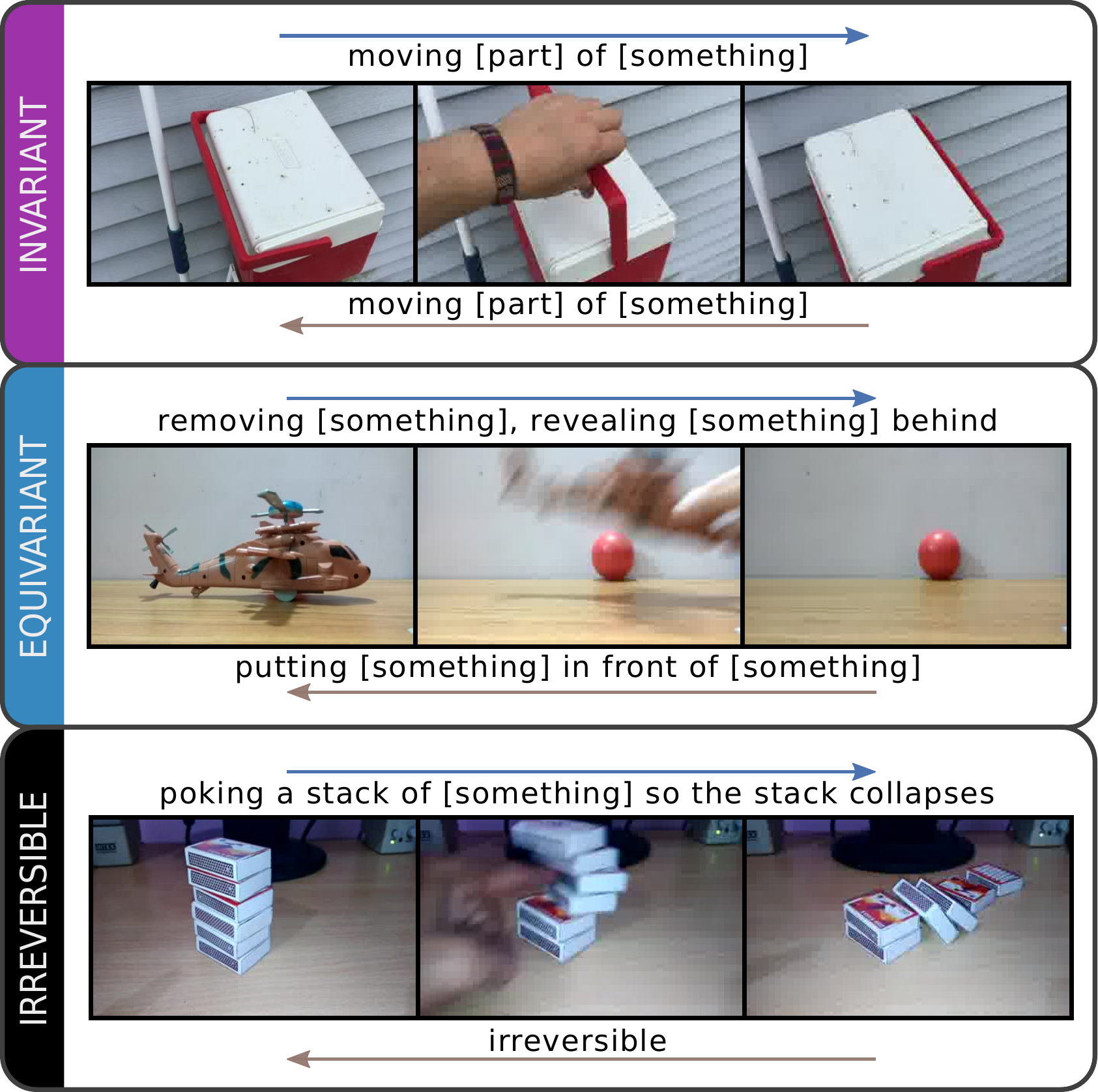}
  \caption{
    When time-reversing a video, \textit{invariant} actions (top) maintain their label,
    \textit{equivariant} actions (middle) exchange labels,
    while some actions (bottom) are \textit{irreversible} producing motions that defy laws of physics.}
  \label{fig:time-reversal-examples}
\end{figure}

\section{Introduction}
\label{sec:introduction}

\noindent Without temporal ordering, individual frames from a video clip of an \class{open jar} action cannot be distinguished from frames of a \class{close jar}.
Tampering with the temporal order, whether through shuffling or reversing the order of frames, has been frequently used to assess the utilisation of temporal signals in action recognition models~\cite{huang2018_WhatMakesVideo,xie2017_RethinkingSpatiotemporalFeature,zhou2017_TemporalRelationalReasoning}.
Recent convolutional models~\cite{wang2017_NonlocalNeuralNetworks,zhou2017_TemporalRelationalReasoning,carreira2017_QuoVadisAction,tran2018_CloserLookSpatiotemporal,xie2017_RethinkingSpatiotemporalFeature}
demonstrate increased robustness by explicitly modelling temporal relations in video.
In a related problem, Arrow of Time (AoT) classification~\cite{Pickup2014_SeeingArrowTime,ghodrati2018_VideoTimeProperties,wei2018_LearningUsingArrow} (the task of determining whether a video is being played forwards or backwards) has been used for pretraining video understanding models.

In this work, we apply the time-reversal video transform on videos to produce new ones that cannot be differentiated from forward-time videos by a human observer.
We validate the realism of these examples through a forced-choice human perception study.
We observe that when time-reversed, reversible videos either maintain their label or undergo a label transformation
(\cref{fig:time-reversal-examples}).
We develop a technique for automatically extracting this label transform for each class from the predictions of a trained classification model.
Next, we apply our findings to other video transforms: horizontal-flipping and the composition of time-reversal with horizontal-flipping.
We then put label transforms to work in zero-shot learning and data augmentation.
Our contributions are summarised as follows:
\begin{itemize}[leftmargin=*,itemsep=-1ex,partopsep=1ex,parsep=1ex]
\item We introduce label-altering video transforms, and identify their corresponding label transforms from model predictions.
\item We evaluate our proposal on two datasets, demonstrating the efficacy of example synthesis for both zero-shot learning and data augmentation.
\item Our zero-shot learning results demonstrate novel opportunities for learning additional classes through video transforms.
  On Something-Something, we learn 16 zero-shot classes \textit{\ie{} without a single example} (out of 174 total classes), and report 46.6\% accuracy compared to 49.5\% with full supervision.
  On Jester, we learn 7 zero-shot classes  (out of 27 total classes), and report 92.4\% accuracy compared to 94.9\% with full supervision.
\end{itemize}

\section{Related work}
\label{sec:related-work}
\noindent In this section, we review relevant works to our proposal related to:
1) the rise in \textit{temporally-sensitive} video recognition models,
2) using time reversal in video
and
3) using video transforms for self-supervision.
To the best of our knowledge, no prior work has investigated label-altering video transforms
for the automatic synthesis of additional labelled training data.

\compactparagraph{Action recognition.}
Action recognition is the task of classifying the action demonstrated in a trimmed video segment.
Classification in early video action recognition datasets~\cite{soomro2012_UCF101Dataset101,kuehne2011_HMDBlargevideo} has been shown to be solvable largely through visual appearance alone~\cite{huang2018_WhatMakesVideo,zhou2017_TemporalRelationalReasoning}. These datasets have been supplanted by larger and more temporally challenging datasets~\cite{kay2017_KineticsHumanAction,goyal2017_SomethingSomethingVideo,Gu2018_AvaVideoDataset,monfort2019_MomentsTimeDataset,damen2018_ScalingEgocentricVision} where this is no longer the case.
This gave rise to papers questioning the ability of both convolutional and recurrent models to capture the temporal order or evolution of the action~\cite{huang2018_WhatMakesVideo,xie2017_RethinkingSpatiotemporalFeature,zhou2017_TemporalRelationalReasoning,Fernando_2015_CVPR,Heidarivincheh2018,dwibedi2018_TemporalReasoningVideos}.
For example, in~\cite{huang2018_WhatMakesVideo} a C3D network trained with hallucinated motion and a single frame from the video is shown to perform comparably to the original video.

Accompanying this evolution has been an increased focus in proposing models that exploit temporal signals in video~\cite{Wang_Transformation,carreira2017_QuoVadisAction,wang2017_NonlocalNeuralNetworks,zhou2017_TemporalRelationalReasoning,tran2018_CloserLookSpatiotemporal,xie2017_RethinkingSpatiotemporalFeature,feichtenhofer2018_SlowFastNetworks}.
In~\cite{Wang_Transformation}, actions are modelled as state transformations, showing improved performance and better generality across actions.
Zhou \etal{}~\cite{zhou2017_TemporalRelationalReasoning} introduce a dedicated layer to correlate the predictions of multiple temporally-ordered video segments, averaging over multiple temporal scales. The model's ability to exploit \textit{time} is tested by shuffling frames in the video.
They report no drop in performance for UCF101, but a clear degradation on Something-Something~\cite{goyal2017_SomethingSomethingVideo} showing the latter is more suitable for learning and evaluating temporal features.

\compactparagraph{Time-reversal in video.}
Time-reversing videos is used for Arrow of Time (AoT) classification~\cite{Pickup2014_SeeingArrowTime,ghodrati2018_VideoTimeProperties,wei2018_LearningUsingArrow}.
First introduced in~\cite{Pickup2014_SeeingArrowTime} and recently revisited in~\cite{wei2018_LearningUsingArrow}, AoT classification is successfully used in self-supervision for pre-training action recognition models.
Of particular relevance to our work is the human perception study of
time-reversed videos on Kinetics by Wei \etal{}~\cite{wei2018_LearningUsingArrow}, showing
humans achieve a 20\% error-rate classifying a video's AoT,
thus demonstrating that dataset subsets contain realistic videos when reversed.

\compactparagraph{Video transforms for self-supervision.}
Video transforms offer a form of self-supervision~\cite{ahsan2019_VideoJigsawUnsupervised,wei2018_LearningUsingArrow,jing2018self,Fernando17,misra2016unsupervised}.
In~\cite{ahsan2019_VideoJigsawUnsupervised}, a video-jigsaw solving task is used for pre-training before fine-tuning for action recognition, and in~\cite{jing2018self} geometric rotation classification is used for pre-training.
In all these works, video transforms are only used in a separate task from which knowledge is transferred to the target task.
The only prior work that has used video transforms for what could be seen as zero-shot learning is~\cite{nair2018_TimeReversalSelfSupervision}.
They utilise time-reversal for training a robot arm to put two blocks together
by observing these blocks exploding apart.

\section{Label-altering video transforms}
\label{sec:non-label-preserving-transforms}
\noindent In this section we introduce label-altering (video) transforms (LATs) and describe how their corresponding class transforms can be determined from predictions of trained models.

\compactparagraph{Introducing LATs.}
Given an oracle video labelling function $f$ and a dataset with videos $V$ and
labels ${Y = \{f(v) \,|\, v \in V\}}$, we aim to learn the parameters of
a model $\hat{f}$ using the videos $V$ and the supervision $Y$.
We define a video transform $\T$ as an operation that takes a video $v \in V$ and transforms it into another video $\hat{v} = \T(v)$ that is a valid input to the trainable model $\hat{f}$.
We restrict our study to video transforms that satisfy the self-inversion property $(\T \circ \T)(v) = v$, and distinguish between two types: \textit{label-preserving} video transforms (LPTs), and \textit{label-altering} video transforms (LATs).
In LPTs, the mapping between a video and its label remains intact

\begin{equation}
  \forall v \in V : f(v) = y \Leftrightarrow f\big(\T(v)\big) = y,
  \label{eq:video-transform-preserving}
\end{equation}
however in LATs, the video transforms which we are interested in, result in a label change such that
\begin{equation}
  \exists v \in V :   f (v) = y \Rightarrow f \big( \T(v) \big) \ne y.
  \label{eq:video-transform-invariant}
\end{equation}
Of all possible LATs, we are interested in ones where the application of the video transform $\T$ to every example of a given class results
in transformed labels belonging to the same class, we call these \textit{class homogeneous} LATs:
\begin{equation}
   \forall \{v, w\} \subset V : f (v) = f (w) \Rightarrow f\big(\T(v)\big) = f\big(\T(w)\big).
\end{equation}
Without class homogeneity, new ground-truth of all transformed videos would be required.
However, when class homogeneity is preserved, class transforms are sufficient to label all transformed videos.
Accordingly, for a class homogeneous LAT, we aim to define the corresponding class transform $\Ty(y)$ for all ${y \in Y}$ where possible.
Given ${V_y = \{v \in V \,|\, f(v) = y\}}$,
we identify three categories of classes:

\begin{enumerate}[leftmargin=*,itemsep=0ex,partopsep=1ex,parsep=1ex]
  \item \textit{Invariant classes}, $Y_i$: classes whose  examples maintain their label after transformation
  \begin{equation}
      Y_i  = \left\{ y \in Y \,|\, \forall v \in V_y  : f\big(\T(v)\big) = y \right\}.
  \end{equation}
  The class transform for invariant classes can thus be defined: $y \in Y_i \Rightarrow \Ty(y) = y$.

  \item \textit{Equivariant}, $Y_e$: classes whose examples change label after transformation
  \begin{equation}
 Y_e = \{ y \in Y \,|\, \exists y' \in Y \; \forall v \in V_y  : f\big(\T(v)\big) = y' \ne y\}.
  \end{equation}
  We thus define $\Ty(y)=y'$, referring to $(y, y')$ as a pair of equivariant classes where $y'$ is the \textit{counterpart} of $y$
  and vice versa.
  Since we desire $\Ty$ to be equivariant to $\T$, we restrict $\Ty$ to be self-invertible, in line with the self-invertible behaviour of $\T$:
  \begin{equation}
      \forall y \in Y_e  : \Ty\big(\Ty(y)\big) = y.
  \end{equation}
    \item \textit{Novel-generating}, $Y_{n}$: these include classes whose
      transformed examples no longer belong to any of the dataset's classes $Y$. We revisit these classes later, using them for zero-shot learning.
  \begin{equation}
       Y_n = \left\{ y \in Y \,|\, y \not\in Y_i \cup Y_e \right\}.
  \end{equation}
\end{enumerate}

\subsection{Discovering class transforms}
\label{sec:class-transforms}

\noindent In order to automatically determine the class transform $\Ty$, we propose a method based on the response of the trained model $\hat{f}$ to all videos from the same dataset transformed by~$\T$.
We first calculate the recall of each class $y$ using the model $\hat{f}$.
We define
\begin{equation}
\hat{V}_y = \{v \in V \,|\, \hat{f}(v) = f(v) = y \},
\end{equation}
and measure the class recall, $\Lambda(y|\hat{f}) = {|\hat{V}_y|}/{|V_y|}$.
If ${\Lambda(y|\hat{f}) \ge \lambda}$ (\ie{} the model performs sufficiently well on that class), the model can be used to establish the class transform $\Tyhat(y)$, assuming minimal noise exists in the dataset labels.
Conversely, if ${\Lambda(y|\hat{f}) < \lambda}$, the class transform cannot be established for $y$ from predictions of the model $\hat{f}$.
We then calculate the proportion of videos in $\hat{V}_y$ that are predicted as $y'$ when $\T$ is applied
\begin{equation}
    \Gamma(y,y'|\hat{f}, \T) = \big|\big\{v \in \hat{V}_y \,|\, \hat{f}\big(\T(v)\big) = y'\big\}\big| / \big|\hat{V}_y\big|,
\end{equation}
and measure affinity between the two classes
\begin{equation}
    \Omega(y, y'|\hat{f}, \T) = \Gamma(y,y'|\hat{f}, \T) \Gamma(y',y|\hat{f}, \T).
\end{equation}
We calculate a candidate target class $y_t$ per class $y$:
\begin{equation}
    y_t = \argmax_{y'}\Omega(y, y' | \hat{f}, \T).
\end{equation}
and introduce a novel target $y_n$ for the class.
Finally, the approximated class transform $\Tyhat$ is:
\begin{equation}
\resizebox{0.88\linewidth}{!}{
  $\hat{\T}_y(y) =
  \begin{cases}
    y &\Omega(y,y) \ge \alpha\\
    y_t &\Omega(y, y_t) \ge \alpha \land \Omega(y,y) < \alpha \land \Omega(y_t, y_t) < \alpha \\
    y_n &\text{otherwise.}
  \end{cases}$}
\end{equation}
where $\alpha$ controls the trade off between extracting invariant and equivariant transforms.

\subsection{Applications of class tranforms}
Next, we describe how class homogenous LATs with their class transforms ($\Ty$) can be used for data augmentation and zero-shot learning.

\compactparagraph{Data augmentation.}
LPTs have long been used for data augmentation and range from the simple, like adjusting the frame rate of a video, to the complex,
like the learnt transformations used in adversarial training~\cite{goodfellow2015_ExplainingHarnessingAdversarial}.
We propose using LATs for augmenting both invariant and equivariant classes through target-conditional data augmentation
\begin{equation}
    V^{\mathrm{aug}}_{y} = V_{y} \cup \{\T(v) \,|\,  v \in V_{y'} \land \T_y(y') = y \in Y \}
\end{equation}

\compactparagraph{Zero-shot learning.}
The novel-generating (NG) classes of $\T$ facilitate zero-shot learning by
synthesising examples of a novel class $y$ as follows:
\begin{equation}
V^{\mathrm{zs}}_y = \{\T(v)\,|\, v \in V_{y'} \land y' \in Y_n \land \Ty(y') = y \}
\end{equation}
The model $\hat{f}$ is trained with synthesised examples
$V^{\mathrm{zs}}_{y}$ of the zero-shot class $y$ and tested on real examples.

\subsection{LAT examples}
\label{sec:for-action}

\noindent We apply the generalisation above on two LATs, as well as their composition (\cref{fig:label-transform-examples}):

\begin{figure}[t]
  \centering
  \includegraphics[width=\linewidth]{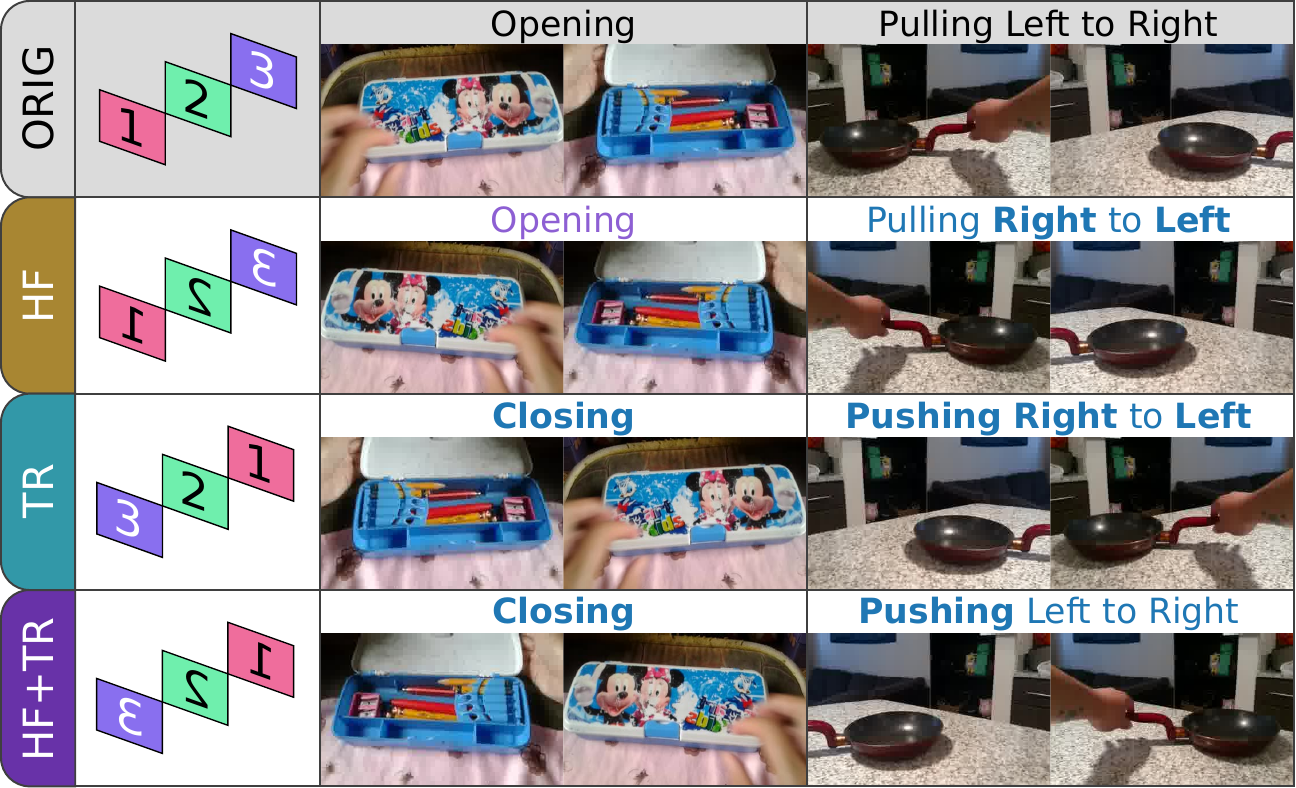}
  \caption{Class transforms for horizontal-flipping, time-reversal, and their composition for two videos from Something-Something (\emph{bold indicates label changes}).}
  \label{fig:label-transform-examples}
\end{figure}

\compactparagraph{Transform 1: horizontal-flipping.}
While in some video datasets, horizontal-flipping is a LPT, it is a LAT when the dataset includes classes with a defining uni-directional horizontal movement, \eg{} \class{swipe right} or \class{rotate clockwise}.

\compactparagraph{Transform 2: time-reversal.}
Unlike horizontal-flipping, time-reversal is a fairly new transform used by the community.
Whilst many classes in action datasets are irreversible, we show that a subset of these maintain realism under time-reversal---an observation that has received little attention.
For example, time reversing an action such as \class{cover} reverses the state change to produce an \class{uncover} action.
We note that many classes invariant under horizontal-flipping become equivariant under time-reversal (\cref{fig:label-transform-examples}).
A number of classes can't be mapped to semantically meaningful classes after the transform as a result of the irreversibility of their examples.

\compactparagraph{What makes a video irreversible?}
We find the realism of reversed videos to be betrayed by \textit{reversal artefacts}, aspects of the scene that would not be possible in a natural world.
Some artefacts are subtle, while others are easy to spot, like a reversed \class{throw} action where the thrown object spontaneously rises from the floor.
We observe two types of reversal artefacts, \textit{physical}, those exhibiting violations of the laws of nature, and \textit{improbable}, those depicting a possible but unlikely scenario.
These are not exclusive, and many reversed actions suffer both types of artefacts, like when uncrumpling a piece of paper.
Examples of physical artefacts include: inverted gravity (\eg{} \class{dropping something}), spontaneous impulses on objects (\eg{} \class{spinning a pen}), and irreversible state changes (\eg{} \class{burning a candle}).
An example of an improbable artefact: taking a plate from the cupboard, drying it, and placing it on the drying rack.

\compactparagraph{Transform 3: horizontal flipping + time reversal.}
We also explore the composition of the two transforms above.
This not only offers new opportunities for data augmentation and zero-shot learning, but also removes some of the biases from the dataset or model.
For example, we note that motion blur affects zero-shot learning when using time-reversal. Combining both transforms removes the model's bias.
Similarly, when a dataset is biased (e.g. more right-handed than left-handed people in our datasets), this composition assists in balancing the dataset.

\section{Datasets and perception study}
\label{sec:datasets}

\noindent To showcase how LATs can be utilised for action recognition, we use two large-scale crowd-sourced datasets.
\textbf{Jester}~\cite{20bn2017_20BNJesterDataset} is a gesture-recognition dataset with 148k videos split into 119k/15k/15k for training/validation/testing
with 27 classes (\eg{} \class{sliding two fingers down}, \class{thumb up}).
\textbf{Something-Something (v2)}~\cite{goyal2017_SomethingSomethingVideo} is an object interaction dataset
containing 221k videos split into
169k/25k/27k for training/val/testing with
174 classes (\eg{}\class{taking something out of something}, \class{tearing something a little bit}).

\compactparagraph{Class transforms.}
We manually define a class transform $\Ty$ for each LAT; this is used as ground truth for both the assessment of the automated discovery of $\Tyhat$, and in evaluating its applications.
We obtain this through inspection of class semantics followed by visual verification.
For horizontal-flipping, we map pairs of classes with defining horizontal motions (e.g. \class{left to right}) to one another and map other classes to themselves.
For time-reversal, we consider what motions and state changes are reversed and
how these interact across classes, then examine reversed examples checking
for reversal artefacts that prevent otherwise reasonable mappings from being defined.

\Cref{tab:label-transform-class-counts} shows the number of classes within each category for the ground truth $\Ty$.
As the table shows, time-reversal results in more equivariant classes than horizontal-flipping.
We find 5 and 28 novel-generating reversible classes in Jester and Something-Something where the transformed label is not part of the label set (\eg{} \class{putting S underneath S} has no counterpart \class{taking S from underneath S}, \textit{S} = something).

\begin{table}[t!]
  \centering
  \begin{adjustbox}{max width=\columnwidth}
  \begin{tabular}{ll rrrr}
    \toprule
                   &                     &              &                & \multicolumn{2}{c}{\# Novel-generating}\\
    \cmidrule(lr){5-6}
    Dataset        & Transform           & \# Invariant & \# Equivariant & Realistic & Unrealistic \\
    \midrule
    Jester         & Horizontal-flip     & 21           & 6              & 0           & 0\\
    Jester         & Time-reverse        & \chl{8}      & \chl{14}       & 5           & 0\\
    \cmidrule{1-6}
    Something      & Horizontal-flip     & 168          & 6              & 0           & 0\\
    Something      & Time-reverse        & \chl{34}     & \chl{32}       & 28          & 80\\
    \bottomrule
  \end{tabular}
  \end{adjustbox}
  \caption{
    Transform class category counts for the ground truth $\Ty$ defined on horizontal-flipping and time-reversal.
    Note the increased number of equivariant and novel-generating classes of time-reversal compared to
    horizontal-flipping.}
  \label{tab:label-transform-class-counts}
\end{table}

\compactparagraph{Arrow of Time: perception study}
Before attempting to use time-reversal as a video transform in our applications, we crowd-sourced a human perception study to confirm the similarity between forward-time and reversed-time examples of our reversible classes.
In \cref{tab:label-transform-class-counts}, we highlight (in blue) the 22 classes from Jester and 66 classes from Something-Something that we deemed time-reversible and on which we conducted this study.

\begin{figure}[t]
  \centering
  \includegraphics[width=\columnwidth]{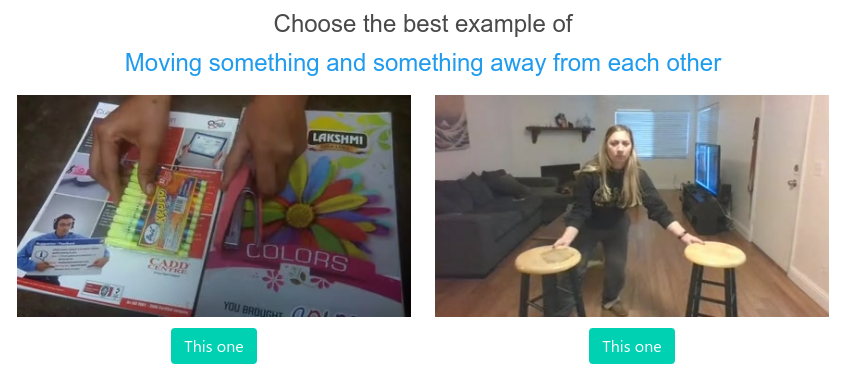}%
  \vspace{-12pt}
  \caption{AMT UI showing an unaltered/time-reversed video.}
  \label{fig:amt-ui}
\end{figure}

Participants were asked to select the better example of two videos in a forced-choice setup (UI shown in \cref{fig:amt-ui}).
They were not given any further instructions of what makes a video a better/worse example of the class, and were not informed that one video was time-reversed.
In each pair, one of the videos was randomly sampled from the training set for that class, while the other was a reversed video sampled from the training set of the label-transformed class.
We randomised the left-right placement of videos.

\begin{figure*}[t]
  \centering
    \includegraphics[width=\textwidth]{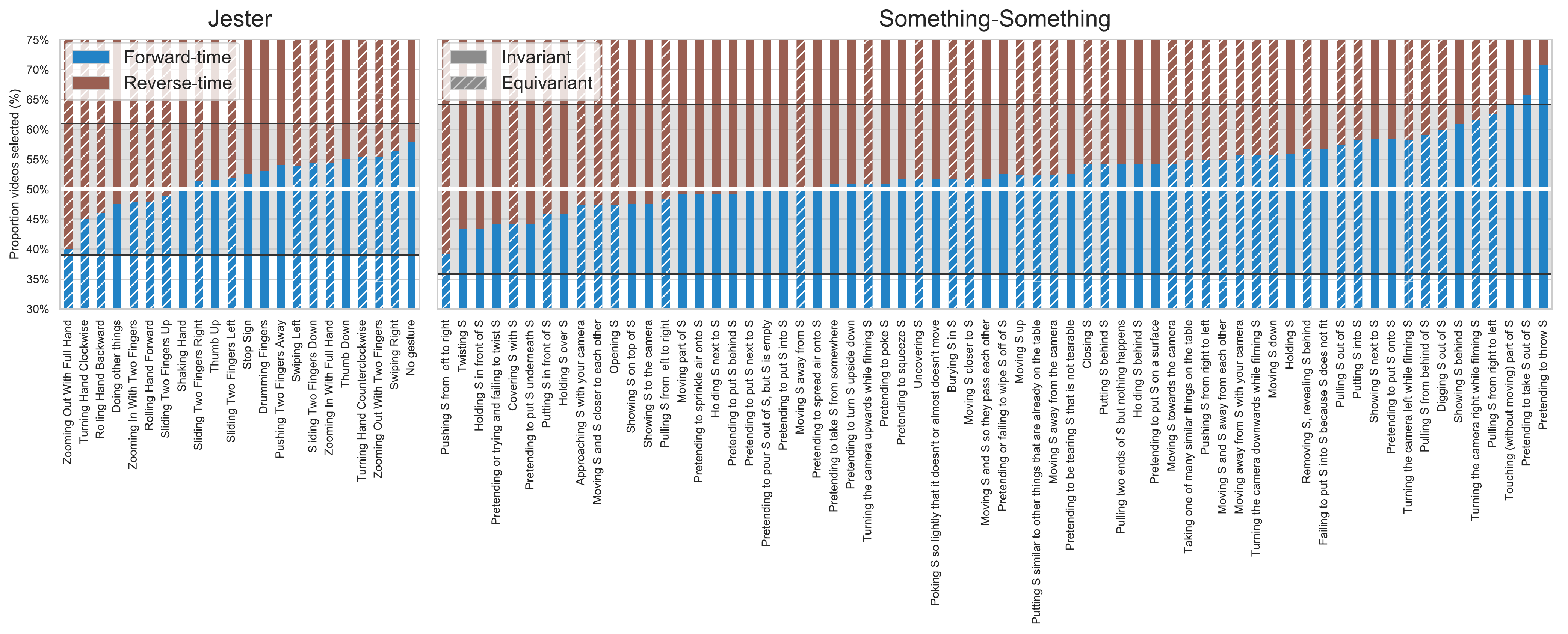}
    \vspace*{-22pt}
    \caption{
      Proportion of forward-time examples selected over reverse-time examples across the 66 reversible classes in Something-Something and 22 in Jester.
    The interval between the dark horizontal lines depicts $50\% \pm 3 \sigma$, within which we consider classes reversible.
  }
  \label{fig:amt-forced-choice}
  \vspace{-5pt}
\end{figure*}

We used Amazon Mechanical Turk (AMT) for the study, testing 20 video pairs in each task.
In $k$ video pairs, the reversed video was replaced with a forward-time video
from an unrelated class as a way to filter out low quality annotations.
We used $k=3$ in Jester and $k=5$ in Something-Something, only accepting submissions that correctly chose 3/3 and 3/5 of these examples respectively.
The bar was set lower for Something-Something due to overlapping classes and occasional low video quality.
In total, 257 individuals annotated 200 videos per class in Jester, and 120 videos per class in Something-Something amounting to 5.8\% and 10.4\% of videos in the reversable class subsets.
To determine which classes are reversible, we model the results for each class as a binomial distribution with $p = 0.5$ approximated by a normal distribution.
We consider classes reversible if their forward-time preference is within $\mu \pm 3\sigma$.
We present the results of this study in \cref{fig:amt-forced-choice}, showing all classes in Jester are within bounds, and only 2 are outside for Something-something (both invariant).
The class with the largest preference for forward-time is \class{pretending to throw something} which exerts asymmetric impulses that participants seem to detect when time-reversed.

Having confirmed that reversed-time examples were sufficiently similar to
forward-time ones in our chosen classes, we move on to using these time-reversed
examples in zero-shot learning and data augmentation.

\section{Experiments and results}
\label{sec:experiments}

\noindent Following a description of implementation details, we examine the
behaviour of the network when exposed to transformed videos, and evaluate our
method to automate class transforms (\cref{sec:experiments:baseline}).
We then present experiments using LATs for zero-shot learning
(\cref{sec:experiments:zeroshot}) and data augmentation (\cref{sec:experiments:data-augmentation}).

\begin{figure*}[ht]
  \centering
  \includegraphics[width=\textwidth]{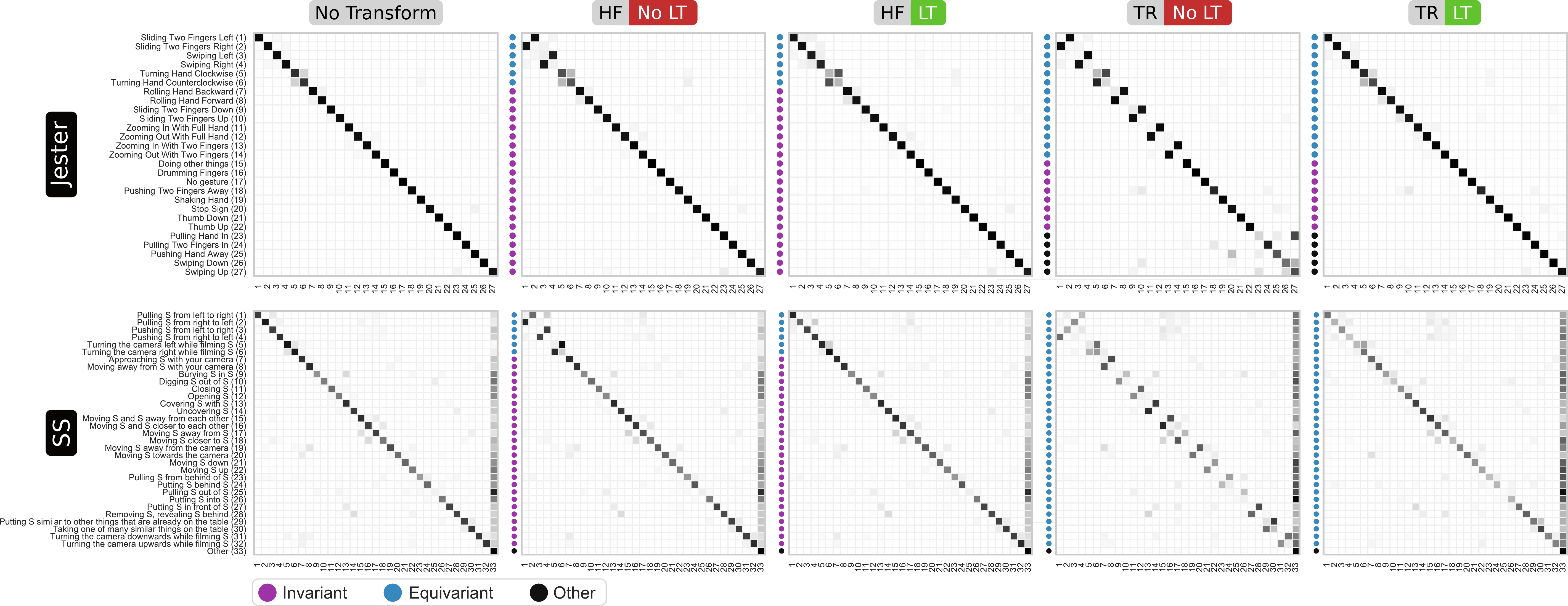}
  \caption{Models trained on forward-time unflipped videos are tested on videos transformed using horizontal flipping or time-reversal, each without or with the correct label transform.
    We only plot the confusion in Something-Something for the 32 time-reversal equivariant classes for clarity. [Best viewed on screen].
  }
  \label{fig:baseline-response-cnf-matrices}
  \vspace{-12pt}
\end{figure*}

\compactparagraph{Implementation details.}
We employ a Temporal Relational Network (TRN)~\cite{zhou2017_TemporalRelationalReasoning} with a batch-normalised Inception (BNInception) backbone~\cite{ioffe2015_BatchNormalizationAccelerating} trained on RGB video due to its temporal-sensitivity, computational efficiency through sparse sampling, and high performance on benchmark datasets (including those we test on).
In TRNs, the input video is split into $n$ segments from which a frame is randomly sampled.
Segment features, extracted by the backbone network, are combined by an MLP to compute temporal relations, followed by class predictions.

We first replicated the validation set results reported by the
authors~\cite{zhou2017_TemporalRelationalReasoning} and assessed the effect of
multi-scale model variant and number of segments before settling on 8-segment single-scale TRN
for our experiments.
We restrict our model evaluations to single center-crops to avoid unintended label transformations introduced by horizontal-flipping.
In all experiments, we train our networks for 100 epochs with an initial learning rate of \num{1e-3} divided by 10 at epochs 40 and 80.
We use a batch size of 80 for Jester and 128 for Something-Something training on 4 GPUs.
All other parameters follow the default values from the TRN GitHub codebase\footnote{\textbf{Data loading issue:} We re-implemented the data loading code but used a different data layout (CTHW)
to the original codebase (TCHW), we actually found this improves results by
$\sim 5\%$ so opted to keep the change. Details on the impact to what the
backbone network is fed with are given in \cref{sec:data-loading-issue}.}.
We report all our results on the validation set of both datasets.

\subsection{Discovering class transforms}
\label{sec:experiments:baseline}

\noindent This first experiment assesses how a model trained
on forward-time videos responds to video transforms, with and without label transformation.
We show the confusion matrices for each LAT in \cref{fig:baseline-response-cnf-matrices}.
For each dataset, we show the baseline performance and the performance after horizontal-flipping or time-reversal without (\textcolor{bad}{red}) and with (\textcolor{good}{green}) label transformation (using the manually defined ground truth label transform).
For easier viewing, we re-order classes so equivariant class pairs are adjacent.
These figures show that equivariant classes are misclassified into their
counterparts without the application of LTs and that when employed, LTs resolve
this misclassification whilst maintaining the correct classification of invariant classes.

One case worthy of note relates to the confusion between \class{turning hand clockwise} and \class{turning hand counterclockwise} in Jester.
Horizontal flipping with LT increases the confusion, which we believe is a result of a population bias towards right-handed people; in a right-handed clockwise hand turn, the back of the hand is shown first then the front, whereas the order is reversed for a left-handed person.

\begin{table}[t!]
  \centering
  \begin{adjustbox}{max width=\columnwidth}
  \footnotesize
  \begin{tabular}{ll rr rrrr}
    \toprule
    Dataset   & Transform & $\lambda$ & $\alpha$ & TP  & FP & FN & TN \\
    \midrule
    Jester    & HF        & 0.90      & 0.80     & 24  & 0  & 2  & 1 \\
    Jester    & TR        & 0.90      & 0.80     & 22  & 1  & 0  & 4 \\
    Something & HF        & 0.04      & 0.09     & 172 & 0  & 2  & 0 \\
    Something & TR        & 0.04      & 0.06     & 59  & 69 & 7  & 39 \\
    \bottomrule
  \end{tabular}
  \end{adjustbox}
  \caption{Evaluation of $\Tyhat$ compared to ground truth $\Ty$.}
  \label{tab:tyhat-evaluation}
\end{table}

Having shown the base model's response to video transforms matches the manually
defined ground truth label transform, we evaluate our method for automatically extracting $\Tyhat$ through the process described in \cref{sec:class-transforms}.
In \cref{tab:tyhat-evaluation}, we report true/false positives/negatives for the $\lambda$, $\alpha$ that maximises the true positive count when treating the extraction of a mapping $y \rightarrow y'$ as a binary classification task.
Note that the optimal $\lambda$, $\alpha$ seem to be independent of the transform, and only different for the dataset/model.
Most class transforms are correctly estimated in Jester for both horizontal-flipping and time-reversal.
For Something-Something, we attribute the larger number of FP due to the models's lower performance
  (49/78\% top-1/5 accuracy) and overlapping classes in the dataset.
Frequently, the established class transforms were reasonable.
For example \class{Moving S away from S} $\leftrightarrow$ \class{Putting S next to S} is a logical mapping, compared to an equally logical ground truth \class{Moving S away from S} $\leftrightarrow$ \class{Moving S closer to S}.

We
investigated the use of NLP for semantically renaming $y$ into its time-reversed
class $y$ by their antonyms, however we found existing lexical databases
lacking. WordNet~\cite{miller1995_WordNetLexicalDatabase} does contain antonym
relations, but these are quite sparse and are missing for common words like
\class{put}, \class{take}, and \class{remove}. Additionally, the antonym
relations that are present are general and don't always embody the time-reversed
class \eg{} \class{move} has the antonym \class{stay}.

In the following sections (\cref{sec:experiments:data-augmentation,sec:experiments:zeroshot}), we report results using the manual ground-truth rather than the discovered ones, avoiding propagating errors into the data augmentation and zero-shot evaluation.
Finally, in \cref{sec:experiments:discovered-transforms}, we test data
augmentation and zero-shot learning using the automatically discovered class transforms.

\subsection{Zero-shot learning}
\label{sec:experiments:zeroshot}

\noindent The novel-generating classes are ideally suited for zero-shot learning, extending the model's recognition abilities to previously-unseen classes.
However, without a test set that includes examples of zero-shot classes, the
model cannot be evaluated.
We instead construct four train/test subsets to evaluate our approach.
We turn pairs of equivariant classes into pairs of novel-generating many-shot
and zero-shot classes.
For each equivariant class pair, we retain the class with the highest
training support as the novel-generating many-shot class and remove all examples
of its counterpart, which then becomes a zero-shot class. The number of zero-shot classes and
corresponding instances synthesised within those classes are listed in \cref{tab:zero-shot-datasets}.

\begin{table}[t]
  \centering
  \begin{adjustbox}{max width=\linewidth}
    \footnotesize
    \begin{tabular}{l rr |l rr}
      \toprule
      Dataset          & \# classes     & \# examples       & Dataset          & \# classes        & \# examples \\
      \midrule
      \jesterzs{HF}    & 3                 & 1387           & \somethingzs{HF} & 3                 & 523 \\
      \jesterzs{TR}    & 7                 & 3450           & \somethingzs{TR} & 16                & 2622 \\
      \bottomrule
    \end{tabular}
  \end{adjustbox}
  \caption{Dataset subset zero-shot class and example counts.
    We still train for all classes in the full dataset, these counts are only for zero-shot classes.}
  \label{tab:zero-shot-datasets}
\end{table}

\begin{table}[t]
  \vspace{-1em}
  \centering
  \small
  \begin{adjustbox}{max width=\linewidth}
  \begin{tabular}{ll rr rr rr}

\toprule

 &                                  & \multicolumn{2}{c}{Zero-shot}                    & \multicolumn{2}{c}{NG many-shot}              & \multicolumn{2}{c}{All classes}                 \\
                                    \cmidrule(lr){3-4}                                 \cmidrule(lr){5-6}                                 \cmidrule(lr){7-8}
 & Supervision                      & Top-1 & Top- 5                                   & Top-1 & Top-5                                    & Top-1 & Top- 5                                  \\
\midrule
\multirow{3}{*}{\rotatebox[origin=c]{90}{\jesterzs{HF}}}
 & \cellcolor{gray} Chance          & \cellcolor{gray} 03.14  & \cellcolor{gray} 14.78 & \cellcolor{gray} 03.17 & \cellcolor{gray} 15.03  & \cellcolor{gray} 04.13 & \cellcolor{gray} 18.69 \\
 & HF                  & 67.92                   & 98.85                  & 91.64                  & 99.50                   & 93.16                  & 99.57                  \\
 & \cellcolor{gray}Full & \cellcolor{gray} 90.34  & \cellcolor{gray} 99.64 & \cellcolor{gray} 90.01 & \cellcolor{gray} 99.65  & \cellcolor{gray} 94.89 & \cellcolor{gray} 99.66 \\

\midrule
\midrule

\multirow{4}{*}{\rotatebox[origin=c]{90}{\jesterzs{TR}}}
 & \cellcolor{gray}Chance           & \cellcolor{gray} 03.35  & \cellcolor{gray} 15.64 & \cellcolor{gray} 03.41 & \cellcolor{gray} 15.99  & \cellcolor{gray} 04.14 & \cellcolor{gray} 18.69 \\
 & TR                     & 78.90                   & 98.70                  & 94.01                  & 99.66                   & 91.99                  & 99.40                  \\
 & TR + HF   & 81.57                   & 99.01                  & 93.04                  & 99.52                   & 92.41                  & 99.46                  \\
 & \cellcolor{gray}Full & \cellcolor{gray} 93.07  & \cellcolor{gray} 99.71 & \cellcolor{gray} 92.61 & \cellcolor{gray} 99.63  & \cellcolor{gray} 94.89 & \cellcolor{gray} 99.66 \\

\midrule
\midrule

\multirow{3}{*}{\rotatebox[origin=c]{90}{\small \somethingzs{HF}}}
 & \cellcolor{gray} Chance          & \cellcolor{gray} 00.76  & \cellcolor{gray} 03.73  & \cellcolor{gray} 00.80 & \cellcolor{gray} 03.99 & \cellcolor{gray} 00.86 & \cellcolor{gray} 04.21 \\
 & HF                  & 71.70                   & 89.29                  & 72.50                  & 90.71                   & 49.38                  & 78.41                  \\
 & \cellcolor{gray} Full & \cellcolor{gray} 77.25  & \cellcolor{gray} 91.20  & \cellcolor{gray} 71.79 & \cellcolor{gray} 89.92 & \cellcolor{gray} 49.45 & \cellcolor{gray} 78.02 \\

\midrule
\midrule

\multirow{4}{*}{\rotatebox[origin=c]{90}{\small \somethingzs{TR}}}
 & \cellcolor{gray} Chance          & \cellcolor{gray} 00.85  & \cellcolor{gray} 04.20  & \cellcolor{gray} 01.10 & \cellcolor{gray} 05.42 & \cellcolor{gray} 00.86 & \cellcolor{gray} 04.21 \\
 & TR                              & 30.93                   & 58.73                   & 61.02                  & 81.42                  & 46.01                  & 75.59                  \\
 & TR + HF    & 39.89                   & 64.80                   & 60.45                  & 80.84                  & 46.56                  & 76.24                  \\
 & \cellcolor{gray} Full & \cellcolor{gray} 62.01  & \cellcolor{gray} 81.88  & \cellcolor{gray} 62.41 & \cellcolor{gray} 83.10 & \cellcolor{gray} 49.45 & \cellcolor{gray} 78.02 \\
\bottomrule
  \end{tabular}
  \end{adjustbox}
  \caption{Zero-shot learning results compared to the upper-bound full-supervision. NG stands for novel-generating.}
  \label{tab:zero-shot}
\end{table}

\begin{figure}[t]
  \centering
  \includegraphics[width=\linewidth]{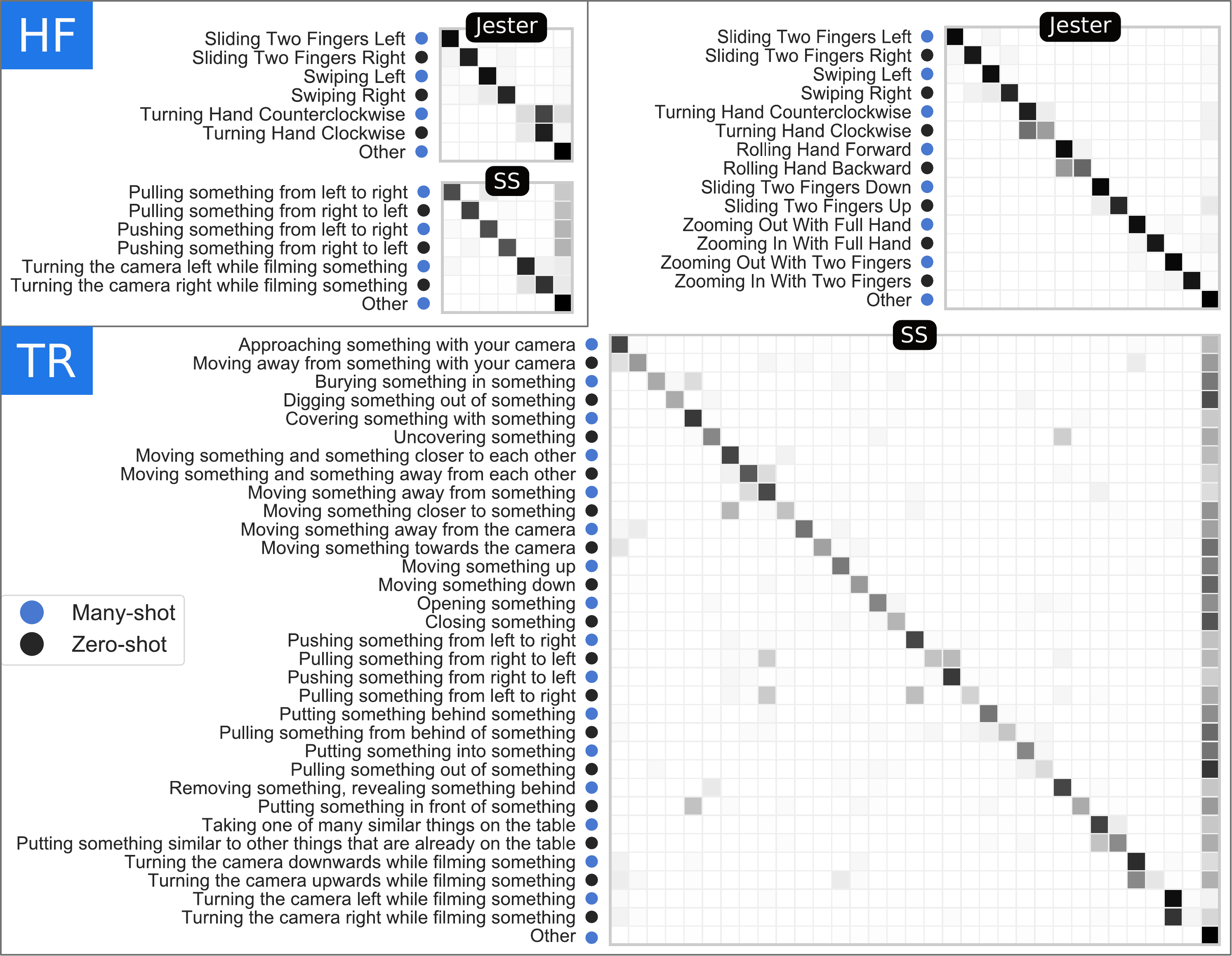}
  \caption{Confusion matrices of the many-shot and zero-shot classes showing
    minimal confusion between zero-shot classes and their many-shot
    counterparts. The final column of each confusion matrix shows confusion
    amongst all other classes not listed.}
  \label{fig:zeroshot-cnf-matrices}
\end{figure}

\begin{figure*}[ht]
  \centering
  \includegraphics[width=\linewidth]{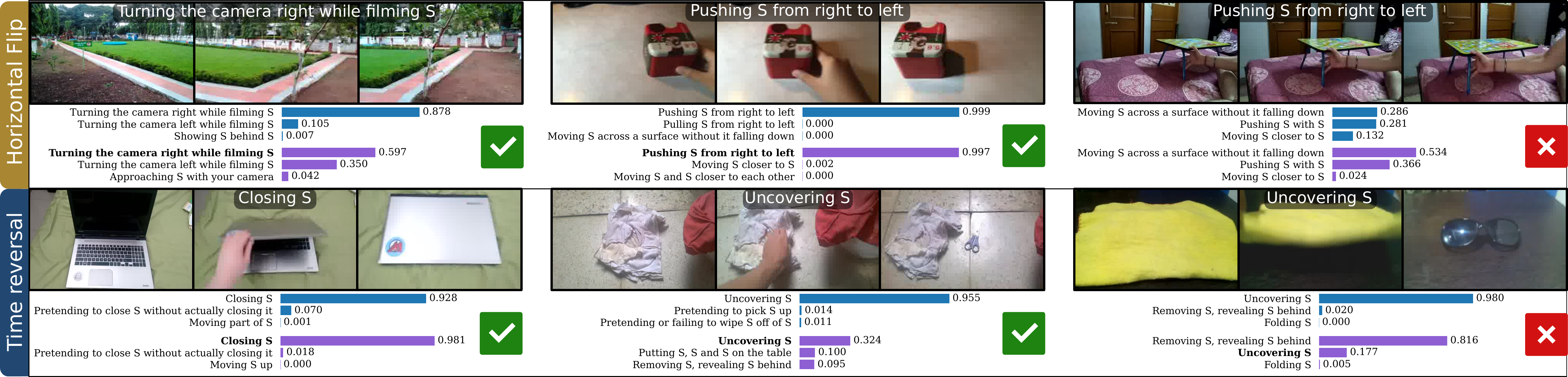}
  \vspace*{-20pt}
  \caption{Sample from SS-HF (top), SS-TR (bottom) comparing the results using
    full supervision vs. zero-shot learning.
    Fully supervised model scores are \textcolor{sns-blue}{blue}, zero-shot model results are \textcolor{sns-purple}{purple} and zero-shot classes are \textbf{bold}.
    [Best viewed on screen].
    }
  \label{fig:qualitative-examples-zero-shot}
\end{figure*}

For each of the four sub-datasets (\jesterzs{HF}, \jesterzs{TR}, \somethingzs{HF}, \somethingzs{TR}), we compare chance (no supervision) as a lower bound to full supervision on all classes as an upper bound.
We present our results in \cref{tab:zero-shot}.
Note that the number of zero-shot and many-shot classes differs per horizontal block.
The results show that training for these zero-shot classes does not affect the
performance of the many-shot classes compared to their full supervision performance.
Over the four subsets, we report an overall drop in top-1 accuracy compared to full supervision of 1.7\%, 2.9\%, 0.1\%, 3.4\%, when dropping all training examples of 11\%, 26\%, 2\% and 9\% classes, respectively.

\Cref{fig:zeroshot-cnf-matrices} shows the confusion matrices for the pairs
of many-shot and zero-shot classes in each
subset. For Jester-HF and Jester-TR, we see good performance, but with the same
confusion between classes \class{turning hand clockwise} and \class{turning hand
  counterclockwise} as in the base model. However, all other classes can be
learnt in the zero-shot setting using horizontal flipping. For SS-HF, zero-shot
classes are distinguishable from their many-shot counterparts. In SS-TR, the
camera movement zero-shot classes: \class{turning the camera upwards/right} have been
confused with their many-shot class counterparts: \class{downwards/left}. This
suggests that the model may be using motion blur in individual frames to
classify the action as the model has never seen upwards/right motion blur
effects.
Overall these confusion matrices show that in the majority of cases, the use of
LATs has resulted in impressive performance on zero-shot classes.

\begin{table*}[ht!]
  \footnotesize
  \centering
  \begin{subtable}[b]{.6\textwidth}
    \centering
    \begin{tabular}[b]{l r rr rr cc}
      \toprule
      &                     & \multicolumn{2}{c}{Zero-shot}   & \multicolumn{2}{c}{NG many-shot}  & \multicolumn{2}{c}{All classes} \\
      \cmidrule(lr){3-4}                \cmidrule(lr){5-6}               \cmidrule(lr){7-8}

      Model       &  $|Y_{\mathrm{zs}}|$  & Top-1 & Top- 5                  & Top-1 & Top-5                  & Top-1 & Top-5 \\
      \midrule
      TR          & 18                  & 32.69 & 57.80                   & 58.70 & 80.44                  & 45.45 \textcolor{bad}{$\blacktriangledown \text{-}0.56$} & 74.77  \textcolor{bad}{$\blacktriangledown \text{-}0.82$}\\
      TR + HF     & 19                  & 38.30 & 61.43                   & 58.46 & 80.68                  & 45.84  \textcolor{bad}{$\blacktriangledown \text{-}0.71$} & 74.92  \textcolor{bad}{$\blacktriangledown \text{-}1.32$}\\
      \bottomrule
    \end{tabular}
  \end{subtable}
  \hspace{0.05\textwidth}
  \begin{subtable}[b]{.3\textwidth}
    \centering
    \begin{tabular}[b]{l cc}
      \toprule
      & \multicolumn{2}{c}{All classes} \\
      \cmidrule(lr){2-3}
      Model                  & Top-1 & Top- 5 \\
      \midrule
      TR                     & 48.52  \textcolor{bad}{$\blacktriangledown \text{-}0.48$} & 77.77 \textcolor{bad}{$\blacktriangledown \text{-}0.14$}\\
      TR + HF                & 49.02 \textcolor{bad}{$\blacktriangledown \text{-}1.25$}& 78.08 \textcolor{bad}{$\blacktriangledown \text{-}0.92$} \\
      \bottomrule
    \end{tabular}
  \end{subtable}
  \caption{Something-Something zero-shot (\textit{left}) and data augmentation
    (\textit{right}) results using extracted class transforms time-reversal (\textit{TR}) or
    horizontal-flipping (\textit{HF}). $|Y_\mathrm{zs}|$~indicates the number of zero-shot classes.}
  \label{tab:extracted-class-transforms-results}
  \vspace{-8pt}
\end{table*}

In \cref{fig:qualitative-examples-zero-shot}, we show qualitative results on six examples from Something-Something.
\textbf{Top Row:} A zero-shot model trained only on left-to-right examples can correctly classify zero-shot right-to-left actions.
The final example shows a case where, although both models incorrectly predict the ground-truth class, their predictions are both reasonable. The zero-shot model has a greater difference between the top-2 scores indicating increased discriminative ability in the model.
\textbf{Bottom Row:}
The time-reversal zero-shot model has been able to learn state inversions like \class{close} (first) and \class{uncover} (second) from time-reversed  examples of \class{open} and \class{cover}.

\begin{table}[ht!]
  \centering
  \begin{adjustbox}{max width=\columnwidth}
  \begin{tabular}{ll c rr rr rr}
    \toprule

    &                                  &       & \multicolumn{2}{c}{All} & \multicolumn{2}{c}{Invariant} & \multicolumn{2}{c}{Equivariant} \\
                                               \cmidrule(lr){4-5}         \cmidrule(lr){6-7}               \cmidrule(l){8-9}
    & Augmentation                     & LT    & Top-1 & Top-5            & Top-1 & Top-5                & Top-1 & Top-5 \\
    \midrule
\multirow{7}{*}{\rotatebox[origin=c]{90}{Jester}}
    & None                             & -     & 94.89 & 99.66            & 95.99 & 99.67                & 90.18 & 99.64 \\
    & HF (invariant only)              & -     & 95.00 & 99.65            & 96.11 & 99.67                & 90.21 & 99.57 \\
    & HF                               & \cross & 94.55 & 99.65            & 96.21 & 99.67                & \cbhl{86.89} & 99.57 \\
    & HF                               & \tick    & 95.01 & 99.67            & 96.25 & 99.67                & 89.71 & 99.68 \\
    \cmidrule(l){2-9}
    & None                             & -     & 94.89 & 99.66            & 97.20 & 99.71                & 92.84 & 99.67 \\
    & TR                               & \tick    & 94.95 & 99.65            & 97.16 & 99.65                & 93.01 & 99.66 \\
    & TR + HF                          & \tick    & 94.68 & 99.61            & 97.06 & 99.73                & 92.55 & 99.56 \\
    \midrule
\multirow{6}{*}{\rotatebox[origin=c]{90}{Something}}
    & None                             & -     & 49.45 & 78.02            & 48.31 & 77.45                & 74.42 & 90.49 \\
    & HF                               & \cross& 49.38 & 78.98            & 49.75 & 78.53                & \cbhl{41.46} & 88.83 \\
    & HF                               & \tick    & 50.26 & 78.94            & 49.20 & 78.36                & 73.50 & 91.78 \\
    \cmidrule(l){2-9}
    & None                             & -     & 49.45 & 78.02            & 36.33 & 70.48                & 62.23 & 82.56 \\
    & TR                               & \tick    & 49.00 & 77.91            & 35.12 & 69.10                & 60.52 & 82.97 \\
    & TR + HF                          & \tick    & 50.27 & 79.00            & 36.95 & 69.85                & 61.23 & 83.52 \\
    \bottomrule
  \end{tabular}
  \end{adjustbox}
  \caption{LAT data augmentation validation set results.
    `LT' stands for label transform, where a hyphen indicates that a LT wouldn't make a difference.
    `Invariant only' refers to applying the data augmentation to the invariant classes solely.
  }
  \label{tab:data-augmentation}
\end{table}

\subsection{Data augmentation}
\label{sec:experiments:data-augmentation}

\noindent We train a model using data augmentation as described in \cref{sec:class-transforms}.
For each video, the transform is applied with a probability of 0.5 along with the corresponding label transform.
This approach results in balancing class support within each equivariant class pair.
In TR+HF, we stack the randomly applied transforms to produce a mixture of
videos with  time-reversal, horizontal-flipping, or their composition.
The results are presented in \cref{tab:data-augmentation}.
Additionally, we include the results of augmenting with horizontal-flipping but without label transformation, as this is a default, yet incorrect, data augmentation technique implemented in TRN and similar video recognition networks. 
This shows a clear drop (highlighted in \cref{tab:data-augmentation}) for equivariant classes.

On Jester, we find the best two configurations to be horizontal-flipping with label transformation and horizontal-flipping of invariant classes only.
Horizontal-flipping with label-transformation improves performance on invariant classes by reducing confusion with equivariant classes.
Time-reversal with label transformation slightly improves performance on equivariant classes.

On Something-Something, We find the combination of time-reversal and horizontal
flipping improves top-1/5 accuracy by 0.8/1.0\%, performing comparably to
horizontal flipping with label transformation alone. Notably, without label
transformation, horizontal flipping results in a model that underperforms the one
trained without augmentation, but with label transformation, the model
outperforms the unaugmented model by 0.8\%.
Note that we used all training examples in addition to transformed ones in this experiment.
Data augmentation for few-shot learning (i.e. by using a subset of the training videos) is left for future work.

\subsection{Using discovered class transforms}
\label{sec:experiments:discovered-transforms}

\noindent Up until this point, we have used manually defined class transforms to report results. This allowed evaluating LATs separately from the
discovery of their class transforms.
We report results on our whole pipeline, on Something-Something, for both TR and HF + TR from discovered class transforms, in
\cref{tab:extracted-class-transforms-results}. The performance is comparable (with a small drop 0.14-1.32\% shown in red) to zero-shot learning
in \cref{tab:zero-shot} and data augmentation in
\cref{tab:data-augmentation}.

\section{Conclusion}
\label{sec:conclusion}

\noindent In this paper, we introduced the notion of label-altering video
transforms, label transforms.
We show example synthesis can be used for zero-shot learning and data augmentation with evaluations on two datasets: Something-Something and Jester.
Future directions involve investigating other label-altering video transforms like video trimming or looping and exploring additional applications of these transforms, \eg{} in few shot learning.
We aim to also investigate learning optimal video transforms to achieve a particular class transform.

{\small
\bibliographystyle{ieee_fullname}
\bibliography{bibliography}

\begin{thebibliography}{10}\itemsep=-1pt

\bibitem{20bn2017_20BNJesterDataset}
20BN.
\newblock The {{20BN}}-{{Jester Dataset}}.
\newblock \url{https://20bn.com/datasets/jester}.
\newblock Online; accessed 2019-02-17.

\bibitem{ahsan2019_VideoJigsawUnsupervised}
Unaiza Ahsan, Rishi Madhok, and Irfan~A. Essa.
\newblock Video jigsaw: Unsupervised learning of spatiotemporal context for
  video action recognition.
\newblock In {\em {IEEE} Winter Conference on Applications of Computer Vision
  (WACV)}, 2019.

\bibitem{carreira2017_QuoVadisAction}
Joao Carreira and Andrew Zisserman.
\newblock Quo vadis, action recognition? a new model and the kinetics dataset.
\newblock In {\em The IEEE Conference on Computer Vision and Pattern
  Recognition (CVPR)}, July 2017.

\bibitem{damen2018_ScalingEgocentricVision}
Dima Damen, Hazel Doughty, Giovanni~Maria Farinella, Sanja Fidler, Antonino
  Furnari, Evangelos Kazakos, Davide Moltisanti, Jonathan Munro, Toby Perrett,
  Will Price, and Michael Wray.
\newblock Scaling egocentric vision: The epic-kitchens dataset.
\newblock In {\em European Conference on Computer Vision (ECCV)}, 2018.

\bibitem{dwibedi2018_TemporalReasoningVideos}
Debidatta Dwibedi, Pierre Sermanet, and Jonathan Tompson.
\newblock Temporal reasoning in videos using convolutional gated recurrent
  units.
\newblock In {\em The IEEE Conference on Computer Vision and Pattern
  Recognition (CVPR) Workshops}, June 2018.

\bibitem{feichtenhofer2018_SlowFastNetworks}
Christoph {Feichtenhofer}, Haoqi {Fan}, Jitendra {Malik}, and Kaiming {He}.
\newblock {SlowFast Networks for Video Recognition}.
\newblock {\em arXiv e-prints}, page arXiv:1812.03982, Dec 2018.

\bibitem{Fernando17}
B. Fernando, H. Bilen, E. Gavves, and S. Gould.
\newblock Self-supervised video representation learning with odd-one-out
  networks.
\newblock In {\em The IEEE Conference on Computer Vision and Pattern
  Recognition (CVPR)}, 2017.

\bibitem{Fernando_2015_CVPR}
Basura Fernando, Efstratios Gavves, Jose~M. Oramas, Amir Ghodrati, and Tinne
  Tuytelaars.
\newblock Modeling video evolution for action recognition.
\newblock In {\em The IEEE Conference on Computer Vision and Pattern
  Recognition (CVPR)}, June 2015.

\bibitem{ghodrati2018_VideoTimeProperties}
Amir Ghodrati, Efstratios Gavves, and Cees Snoek.
\newblock Video {{Time}}: {{Properties}}, {{Encoders}} and {{Evaluation}}.
\newblock In {\em British Machine Vision Conference (BMVC)}, page 160, 2018.

\bibitem{goodfellow2015_ExplainingHarnessingAdversarial}
Ian Goodfellow, Jonathon Shlens, and Christian Szegedy.
\newblock Explaining and harnessing adversarial examples.
\newblock In {\em International Conference on Learning Representations (ICLR)},
  2015.

\bibitem{goyal2017_SomethingSomethingVideo}
Raghav Goyal, Samira Ebrahimi~Kahou, Vincent Michalski, Joanna Materzynska,
  Susanne Westphal, Heuna Kim, Valentin Haenel, Ingo Fruend, Peter Yianilos,
  Moritz Mueller-Freitag, Florian Hoppe, Christian Thurau, Ingo Bax, and Roland
  Memisevic.
\newblock The ``{{Something Something}}'' {{Video Database}} for {{Learning}}
  and {{Evaluating Visual Common Sense}}.
\newblock In {\em The {{IEEE International Conference}} on {{Computer Vision}}
  ({{ICCV}})}, 2017.

\bibitem{Gu2018_AvaVideoDataset}
Chunhui Gu, Chen Sun, David~A. Ross, Carl Vondrick, Caroline Pantofaru, Yeqing
  Li, Sudheendra Vijayanarasimhan, George Toderici, Susanna Ricco, Rahul
  Sukthankar, Cordelia Schmid, and Jitendra Malik.
\newblock Ava: A video dataset of spatio-temporally localized atomic visual
  actions.
\newblock In {\em The IEEE Conference on Computer Vision and Pattern
  Recognition (CVPR)}, June 2018.

\bibitem{Heidarivincheh2018}
Farnoosh Heidarivincheh, Majid Mirmehdi, and Dima Damen.
\newblock Action completion: A temporal model for moment detection.
\newblock In {\em British Machine Vision Conference (BMVC)}, 2018.

\bibitem{huang2018_WhatMakesVideo}
De-An Huang, Vignesh Ramanathan, Dhruv Mahajan, Lorenzo Torresani, Manohar
  Paluri, Li Fei-Fei, and Juan Carlos~Niebles.
\newblock What makes a video a video: Analyzing temporal information in video
  understanding models and datasets.
\newblock In {\em The IEEE Conference on Computer Vision and Pattern
  Recognition (CVPR)}, June 2018.

\bibitem{ioffe2015_BatchNormalizationAccelerating}
Sergey Ioffe and Christian Szegedy.
\newblock Batch normalization: Accelerating deep network training by reducing
  internal covariate shift.
\newblock In Francis Bach and David Blei, editors, {\em Proceedings of the 32nd
  International Conference on Machine Learning ({ICML})}. PMLR, 2015.

\bibitem{jing2018self}
Longlong {Jing}, Xiaodong {Yang}, Jingen {Liu}, and Yingli {Tian}.
\newblock {Self-Supervised Spatiotemporal Feature Learning via Video Rotation
  Prediction}.
\newblock {\em arXiv e-prints}, page arXiv:1811.11387, Nov 2018.

\bibitem{kay2017_KineticsHumanAction}
Will {Kay}, Joao {Carreira}, Karen {Simonyan}, Brian {Zhang}, Chloe {Hillier},
  Sudheendra {Vijayanarasimhan}, Fabio {Viola}, Tim {Green}, Trevor {Back},
  Paul {Natsev}, Mustafa {Suleyman}, and Andrew {Zisserman}.
\newblock {The Kinetics Human Action Video Dataset}.
\newblock {\em arXiv e-prints}, page arXiv:1705.06950, May 2017.

\bibitem{kuehne2011_HMDBlargevideo}
Hilde Kuehne, Hueihan Jhuang, Estibaliz Garrote, Tomaso Poggio, and Thomas
  Serre.
\newblock {{HMDB}}: {{A}} large video database for human motion recognition.
\newblock In {\em The {{IEEE International Conference}} on {{Computer Vision}}
  ({{ICCV}})}, 2011.

\bibitem{miller1995_WordNetLexicalDatabase}
George~A. Miller.
\newblock Wordnet: A lexical database for english.
\newblock {\em Commun. ACM}, 38(11):39--41, Nov. 1995.

\bibitem{misra2016unsupervised}
Ishan Misra, C.~Lawrence Zitnick, and Martial Hebert.
\newblock {Shuffle and Learn: Unsupervised Learning using Temporal Order
  Verification}.
\newblock In {\em The European Conference on Computer Vision (ECCV)}, 2016.

\bibitem{monfort2019_MomentsTimeDataset}
Mathew Monfort, Alex Andonian, Bolei Zhou, Kandan Ramakrishnan, Sarah~Adel
  Bargal, Tom Yan, Lisa Brown, Quanfu Fan, Dan Gutfruend, Carl Vondrick, et~al.
\newblock Moments in time dataset: one million videos for event understanding.
\newblock {\em IEEE Transactions on Pattern Analysis and Machine Intelligence
  (TPAMI)}, 2019.

\bibitem{nair2018_TimeReversalSelfSupervision}
Suraj {Nair}, Mohammad {Babaeizadeh}, Chelsea {Finn}, Sergey {Levine}, and
  Vikash {Kumar}.
\newblock {Time Reversal as Self-Supervision}.
\newblock {\em arXiv e-prints}, page arXiv:1810.01128, Oct 2018.

\bibitem{Pickup2014_SeeingArrowTime}
Lyndsey~C. Pickup, Zheng Pan, Donglai Wei, YiChang Shih, Changshui Zhang,
  Andrew Zisserman, Bernhard Scholkopf, and William~T. Freeman.
\newblock Seeing the arrow of time.
\newblock In {\em The IEEE Conference on Computer Vision and Pattern
  Recognition (CVPR)}, June 2014.

\bibitem{soomro2012_UCF101Dataset101}
Khurram {Soomro}, Amir {Roshan Zamir}, and Mubarak {Shah}.
\newblock {UCF101: A Dataset of 101 Human Actions Classes From Videos in The
  Wild}.
\newblock {\em arXiv e-prints}, page arXiv:1212.0402, Dec 2012.

\bibitem{tran2018_CloserLookSpatiotemporal}
Du Tran, Heng Wang, Lorenzo Torresani, Jamie Ray, Yann LeCun, and Manohar
  Paluri.
\newblock A closer look at spatiotemporal convolutions for action recognition.
\newblock In {\em The IEEE Conference on Computer Vision and Pattern
  Recognition (CVPR)}, June 2018.

\bibitem{Wang_Transformation}
Xiaolong Wang, Ali Farhadi, and Abhinav Gupta.
\newblock Actions {\textasciitilde} transformations.
\newblock In {\em The IEEE Conference on Computer Vision and Pattern
  Recognition (CVPR)}, 2016.

\bibitem{wang2017_NonlocalNeuralNetworks}
Xiaolong Wang, Ross Girshick, Abhinav Gupta, and Kaiming He.
\newblock Non-local neural networks.
\newblock In {\em The IEEE Conference on Computer Vision and Pattern
  Recognition (CVPR)}, 2018.

\bibitem{wei2018_LearningUsingArrow}
Donglai Wei, Joseph~J. Lim, Andrew Zisserman, and William~T. Freeman.
\newblock Learning and using the arrow of time.
\newblock In {\em The IEEE Conference on Computer Vision and Pattern
  Recognition (CVPR)}, June 2018.

\bibitem{xie2017_RethinkingSpatiotemporalFeature}
Saining Xie, Chen Sun, Jonathan Huang, Zhuowen Tu, and Kevin Murphy.
\newblock Rethinking spatiotemporal feature learning: Speed-accuracy trade-offs
  in video classification.
\newblock In {\em The European Conference on Computer Vision (ECCV)}, September
  2018.

\bibitem{zhou2017_TemporalRelationalReasoning}
Bolei Zhou, Alex Andonian, Aude Oliva, and Antonio Torralba.
\newblock Temporal relational reasoning in videos.
\newblock In {\em The European Conference on Computer Vision (ECCV)}, September
  2018.

\end{thebibliography}
}
\begin{figure*}[t]
  \centering
  \includegraphics[width=\textwidth]{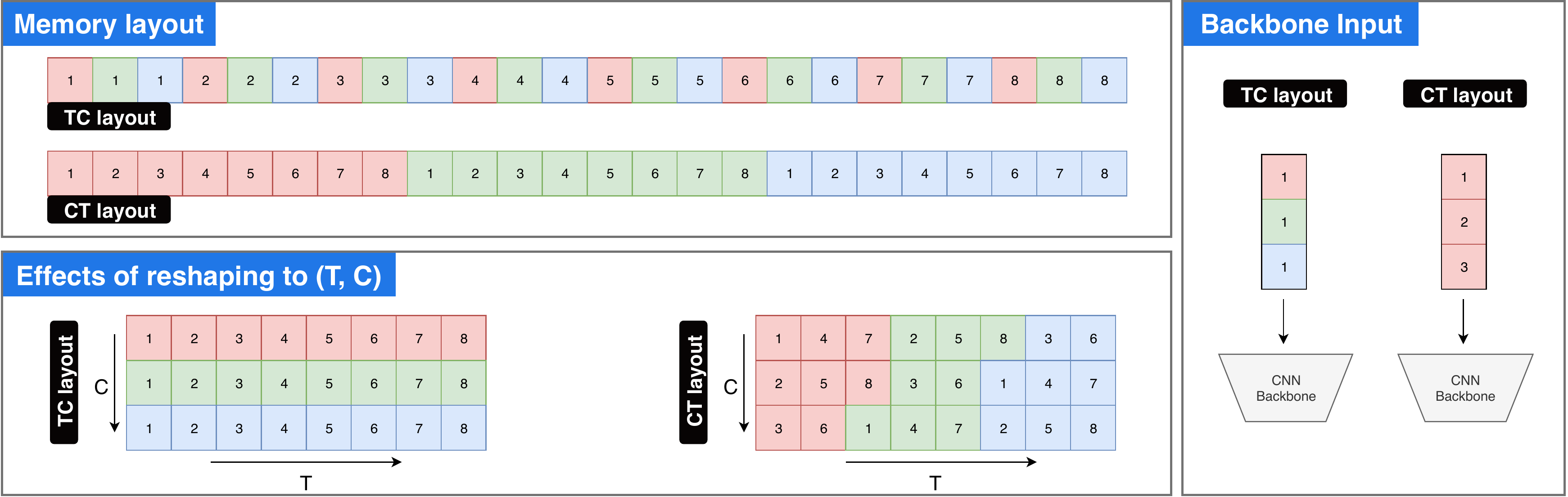}
  \caption{\textit{Top}: The layout of TCHW and CTHW data in memory. The numbers
    indicate the frame and the colour, the channel (RGB). \textit{Bottom}: The
    resulting layout of elements after reshaping TCHW (left) or CTHW (right) data
    into a tensor of shape $(T, C, H, W)$ \textit{without any transposition}. \textit{Right}: The input to the 2D
    CNN backbone when slicing columns of the $(T, C, H, W)$ tensor. Note we have
    only visualised the time (T) and channel (C) dimensions for clarity since
    the remaining dimensions data-layout is unchanged.}
  \label{fig:cthw-tchw-issue}
\end{figure*}

\newpage
\appendix
\section{Data loading issue}
\label{sec:data-loading-issue}
After the camera-ready submission of our ICCVW paper, we found a set of mismatched assumptions in our
codebase: the data loading code loaded videos with CTHW (channels, time, height,
width) layout tensors, but the TRN implementation expected TCHW layout tensors.
The TRN implementation uses a reshape operation on its input to squash the time
dimension into the batch dimension for propagating all frames in the batch
through the 2D CNN backbone. The effects of this mismatch between the expected
and actual data-layout on the input to the 2D CNN backbone are visualised in
\cref{fig:cthw-tchw-issue}.

To quantify the impact of this error, we re-ran a set of experiments focusing on
those evaluating the use of time-reversal on Something-Something, the results of which are presented
for zero-shot and data-augmentation in \cref{tab:zero-shot-data-layout-changes} and \cref{tab:data-augmentation-data-layout-changes}.
Surprisingly, we found that feeding
data in the \textit{incorrect} CTHW format led to \textit{improved} performance
across all our experiments, including standard experiments that do not employ any video transforms. The CTHW layout improved overall recognition results on Something-Something validation set from 44.95\% to 49.45\% (4.5\% improvement). We posit this is due to the backbone being able to
exploit temporal signals as temporal information is fed into the 2D CNN backbone
unlike with the TCHW format tensor.

This issue has not materially impacted the conclusions of the paper, and since
we achieve improved performance with the CTHW data layout, we have opted to
keep the current set of experimental results.

\begin{table*}[t]
  \centering
  \begin{tabular}{l rrr rrr rrr}
    \toprule
                  & \multicolumn{3}{c}{All}    & \multicolumn{3}{c}{Many-shot} & \multicolumn{3}{c}{Zero-shot} \\
                 \cmidrule(l){2-4}            \cmidrule(lr){5-7}                \cmidrule(r){8-10}
    Supervision   & TCHW  & CTHW  & $\delta$       & TCHW  & CTHW  & $\delta$            & TCHW  & CTHW  & $\delta$ \\
    \midrule
    Full          & 44.95 & 49.45 & \textcolor{good}{+4.50}      & 54.13 & 62.41 & \textcolor{good}{+8.28}           & 51.72 & 62.01 & \textcolor{good}{+10.29} \\
    Time-reversal & 41.65 & 46.01 & \textcolor{good}{+4.36}      & 51.27 & 61.02 & \textcolor{good}{+9.75}           & 24.29 & 30.93 & \textcolor{good}{+6.64} \\
    \bottomrule
  \end{tabular}
  \caption{Comparison of top-1 accuracy on Something-Something zero-shot experiments when training
    with data laid out in TCHW or CTHW.}
  \label{tab:zero-shot-data-layout-changes}
\end{table*}

\begin{table*}[t]
  \centering
  \begin{tabular}{l rrr rrr rrr}
    \toprule
                 & \multicolumn{3}{c}{All}    & \multicolumn{3}{c}{Equivariant} & \multicolumn{3}{c}{Invariant} \\
                 \cmidrule(lr){2-4}            \cmidrule(lr){5-7}                \cmidrule(lr){8-10}
    Augmentation & TCHW  & CTHW  & $\delta$       & TCHW  & CTHW  & $\delta$            & TCHW  & CTHW  & $\delta$ \\
    \midrule
    None         & 44.95 & 49.45 & \textcolor{good}{+4.50}      & 53.06 & 62.23 & \textcolor{good}{+9.17}           & 36.09 & 36.33 & \textcolor{good}{+0.24} \\
    TR           & 44.62 & 49.00 & \textcolor{good}{+4.38}      & 51.31 & 60.52 & \textcolor{good}{+9.21}           & 33.63 & 35.12 & \textcolor{good}{+1.49} \\
    \bottomrule
  \end{tabular}
  \caption{Comparison of top-1 accuracy on Something-Something data-augmentation experiments when
    training with data laid out in TCHW or CTHW.}
  \label{tab:data-augmentation-data-layout-changes}
\end{table*}

\end{document}